\documentclass{article}
\usepackage{ijcai26}

\usepackage{times}
\usepackage{soul}
\usepackage{url}
\usepackage[hidelinks]{hyperref}
\usepackage[utf8]{inputenc}
\usepackage[small]{caption}
\usepackage{graphicx}
\usepackage{amsmath}
\usepackage{amsthm}
\usepackage{booktabs}
\usepackage{algorithm}
\usepackage{algorithmic}
\usepackage[switch]{lineno}
\usepackage{amsmath}
\usepackage{amssymb}
\usepackage{multirow}
\usepackage{array}
\usepackage{booktabs}
\usepackage{graphicx}
\usepackage{float}
\usepackage{hyperref}

\title{CHoE: Cross-Domain Heterogeneous Graph Prompt Learning via Structure-Conditioned Experts}

\author{
Peiyuan Li$^{1\dagger}$ \and
Yongqi Huang$^{1\dagger}$ \and
Jitao Zhao$^{1}$\and
Dongxiao He$^{1*}$ \and
Di Jin$^{1}$\And
Weixiong Zhang$^{2}$
\affiliations
$^{1}$School of Computer Science and Technology, Tianjin University, Tianjin, China\\
$^{2}$Department of Health Technology and Informatics, and Department of Data Science and Artificial Intelligence, The Hong Kong Polytechnic University, Kowloon, Hong Kong\\
\emails
\{lipeiyuan04, yqhuang, zjtao, hedongxiao, jindi\}@tju.edu.cn,
weixiong.zhang@polyu.edu.hk
}

\begin{document}

\maketitle
{
  \renewcommand{\thefootnote}{}
  \footnotetext{$^{\dagger}$These authors contributed equally.}
  \footnotetext{$^{*}$Corresponding author.}
  \footnotetext{Code is avaliable at: \href{https://github.com/hedongxiao-tju/CHoE}{https://github.com/hedongxiao-tju/CHoE}}
}

\begin{abstract}

Heterogeneous Graph Prompt Learning (HGPL) has emerged as a promising paradigm for bridging the gap between the objectives of pre-training foundation models and their downstream applications in heterogeneous graph settings. However, existing HGPL methods are primarily designed for in-domain scenarios, whereas real-world deployments often span multiple domains, and the data used for pre-training and downstream tasks may originate from different distributions. Consequently, the applicability of current HGPL approaches is limited to in-domain settings, and their performance typically degrades when application domains shift. To address this serious limitation, we develop CHoE, a cross-domain HGPL method built upon an expert network. During pre-training, we introduce and train structure-conditioned experts, and during prompt tuning, we adopt a structure-aware expert routing and load balancing mechanism to select structurally compatible experts for each meta-path view. In addition, we design a prompt-based semantic fusion module to integrate representations across multiple views for downstream prediction. Extensive experiments show that CHoE consistently improves performance in few-shot cross-domain applications, outperforming all baseline approaches.

\end{abstract}

\section{Introduction}
\label{sec:introduction}
Heterogeneous Graphs (HGs) are ubiquitous across nearly all application domains, including biomedical~\cite{BioMedical} networks in biological and medical applications and bibliographic~\cite{bibliographic} networks in scientific research. In recent years, Heterogeneous Graph Neural Networks (HGNNs) have attracted significant attention for modeling and analyzing HGs. Most HGNN methods rely heavily on large amounts of labeled data, which is costly to obtain. To mitigate label dependency, ``the pre-training and then fine-tuning'' paradigm provides an effective alternative by using unlabeled data to learn and adapt general knowledge to specific downstream tasks. This paradigm reduces dependence on labeled data while maintaining transferability across diverse downstream applications~\cite{bibliographic}.

In practice, adapting a pre-trained model to accommodate diverse downstream conditions requires efficient, robust mechanisms beyond task-specific full fine-tuning~\cite{MUG}. Inspired by the success of prompt learning in natural language~\cite{prompt_for_nlp} and computer vision~\cite{prompt_for_cv}, recent studies have attempted to extend this paradigm to graphs. Graph Prompt Learning (GPL)~\cite{GPL_survey} has emerged as a powerful and versatile approach. By constructing flexible, lightweight prompting modules for specific downstream tasks, GPL aligns pre-trained models with downstream tasks and alleviates negative transfer~\cite{Uniprompt}. Most existing GPL methods fix the parameters of pre-trained models and only fine-tune lightweight prompt modules, enabling efficient adaptation of pre-trained models to downstream tasks. Some studies have explored GPL methods in heterogeneous graphs. HGPrompt~\cite{HGPrompt} decomposes a heterogeneous graph into multiple homogeneous subgraphs and introduces unified-feature and subgraph-specific prompts, enabling models to adapt to downstream tasks. HetGPT~\cite{HetGpt} further enriches prompt design by incorporating virtual-class prompts and introducing node-specific feature prompts. HetGPT also employs a multi-view neighborhood aggregation mechanism to model semantic and structural information in heterogeneous graphs.

\begin{figure}[t]
    \centering
    \includegraphics[width=1.0\linewidth]{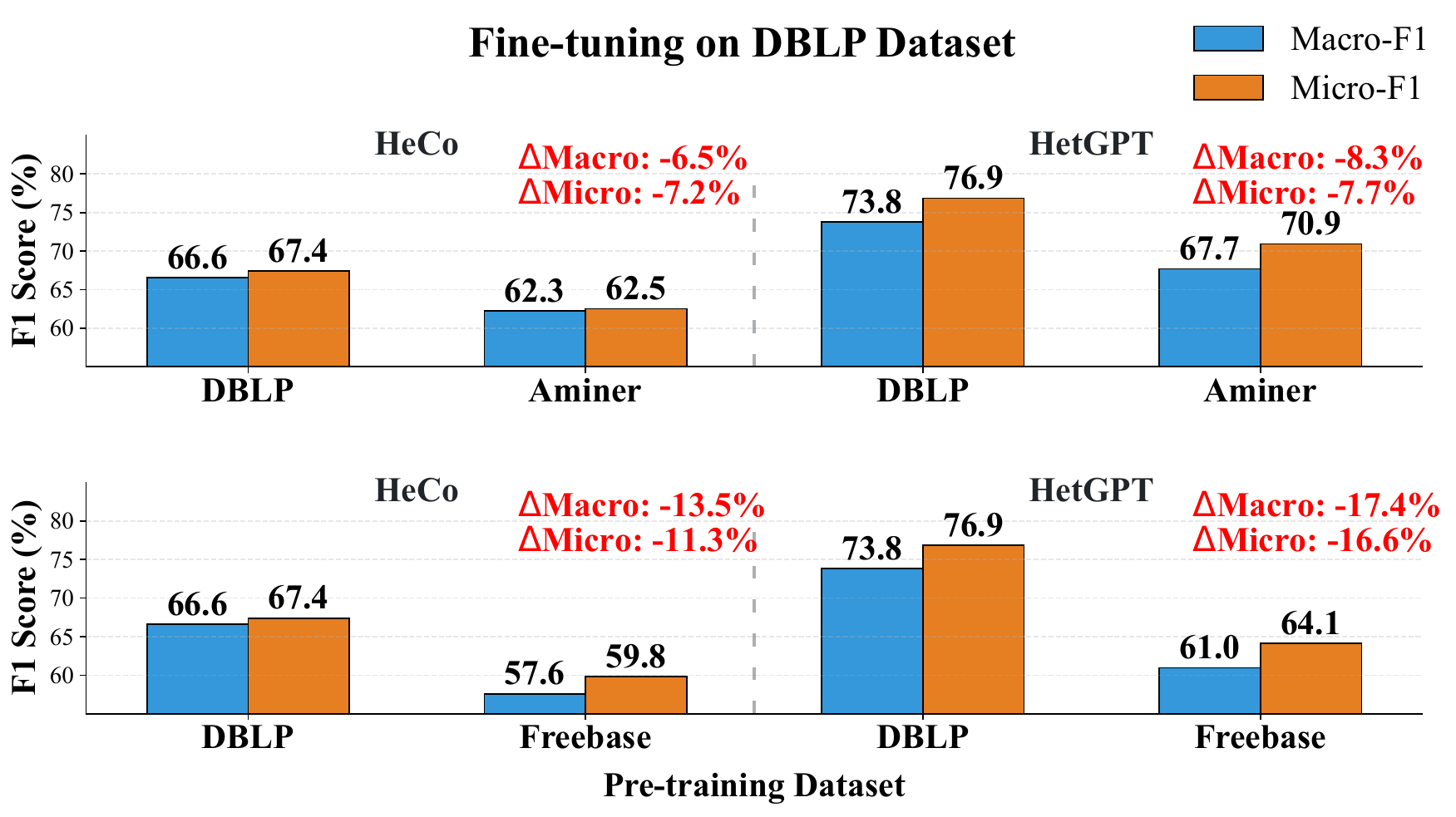}
    \caption{Motivation Experiments. We present the performance of HeCo and HetGPT when pre-trained on Aminer and Freebase and fine-tuned on the DBLP dataset. The percentage drops ($\Delta$) reflect the performance degradation caused by domain shifts compared with in-domain settings (i.e., pre-training and fine-tuning on DBLP).}
    \label{fig:motivation}
\end{figure}

However, despite recent advancements, existing HGPL methods are typically designed for in-domain scenarios. In real-world applications, models are often required to operate across domains, where source and target domains follow distinct distributions~\cite{GPH2,TFSGFM}.
For example, a model may be pre-trained for biomedical networks but is applied to downstream applications on chemicals. The applicability of these HGPL methods remains limited for cross-domain scenarios. Changes in application domains typically degrade the performance of HGPL methods. As shown in Figure~\ref{fig:motivation}, existing HGPL methods exhibit reduced performance in cross-domain settings, and carefully designed prompt modules are less competitive than fully fine-tuning the pre-trained model. Therefore, addressing the performance degradation problem in cross-domain scenarios is an urgent issue in heterogeneous graph learning.

Effective cross-domain learning on heterogeneous graphs faces two key challenges. The first is making a pre-trained model compatible with downstream graphs from different domains. In such settings, both the types and counts of meta-paths can vary. Because existing methods are tightly coupled to pre-defined meta-paths, their pre-trained representations transfer poorly to new domains. The second challenge is enabling knowledge transfer under semantic and structural distribution shifts. Semantic and structural patterns often differ significantly across domains. Even simple adaptations that render current methods nominally cross-domain tend to fail under these shifts, resulting in noticeable performance degradation. Therefore, there is a clear need for a cross-domain framework that keeps pre-trained models compatible with diverse downstream graphs while explicitly bridging semantic and structural discrepancies between domains.

To address these challenges, we propose CHoE, a cross-domain HGPL method built upon an expert network perspective. In the pre-training stage, we design a generative self-supervised learning framework that reconstructs masked node features and edges, encouraging the model to learn universal semantic and structural knowledge from the source domain. In the prompt-tuning stage, we introduce structure-conditioned experts learned during pre-training, allowing each meta-path view in the downstream graph to adaptively select and combine expert representations. Specifically, we develop a structure-aware expert routing mechanism that evaluates experts according to their ability to model the downstream graph structure and selects those that are structurally compatible, together with a load balancing mechanism that uses accumulated routing statistics from previous epochs to promote balanced expert usage and prevent biased expert selection. At the semantic level, we propose a prompt-based semantic fusion module that introduces meta-path-specific prompts to enable adaptive fusion of representations from different semantic views, thereby achieving more robust and efficient knowledge transfer across domains. In summary, our main contributions are as follows:
\begin{itemize}
 
    \item We identify an under-explored but practically important cross-domain scenario for heterogeneous graph and show that existing HGPL methods degrade under semantic and structural shifts.

    \item We propose CHoE, a cross-domain HGPL method based on structure-conditioned experts. By transferring a pre-trained expert pool and enabling structure-aware routing and load balancing, CHoE mitigates cross-domain structural discrepancies. Moreover, a prompt-based semantic fusion module is introduced to achieve cross-domain semantic adaptation.

    \item We conduct extensive few-shot experiments on four heterogeneous graph datasets. The results consistently demonstrate the effectiveness and robustness of CHoE in both cross-domain and few-shot settings.
\end{itemize}

\section{Related Work}

\textbf{Heterogeneous graph pre-training.} Graph pre-training has attracted growing attention in recent years, and numerous methods for heterogeneous graphs have been proposed. Among them, DMGI~\cite{DMGI} and HDMI~\cite{HDMI} extend DGI~\cite{DGI} to heterogeneous settings by maximizing mutual information between node representations and graph-level summaries within each meta-path view and then aggregating multi-view representations into a unified embedding. Inspired by graph contrastive learning, HeCo~\cite{HeCo} employs a co-contrastive framework that contrasts local and higher-order views, whereas HGCML~\cite{HGCML} performs multi-view contrastive learning by constructing several meta-path-based views. HGMAE~\cite{HGMAE} adopts a generative approach that reconstructs masked node features and masked edges. In contrast to these meta-path-dependent approaches, HERO~\cite{HERO} dispenses with pre-defined meta-paths by modeling homophily through a self-expression matrix and capturing heterogeneity via a type-aware attention-based aggregation of node representations

\textbf{Heterogeneous graph prompt learning.} Heterogeneous graph pre-training methods have shown strong representation learning capabilities but often suffer from negative transfer when adapted to diverse downstream tasks. HGPL has thus emerged as a new paradigm. HGPrompt~\cite{HGPrompt} decomposes a heterogeneous graph into multiple homogeneous subgraphs and introduces unified-feature and subgraph-specific prompts, while HetGPT~\cite{HetGpt} further enriches the prompt space by incorporating virtual-class prompts and node-specific feature prompts, together with a multi-view neighborhood aggregation mechanism to capture complex neighborhood. HiGPT~\cite{HiGPT} leverages context-parameterized heterogeneity projectors and large language models to generate node representations and applies instruction tuning to better align with downstream tasks. Distinct from prior work that primarily emphasizes prompt design, GPAWP~\cite{GPAWP} instead investigates how prompts function in practice, proposing a prompt-importance evaluation mechanism to quantify their contributions and prune ineffective or negatively impactful prompts.

Despite their effectiveness, existing methods remain limited to in-domain settings and have not yet fully achieved cross-domain transfer, which this paper aims to address.

\section{Preliminary}

\textbf{Definition 3.1 Heterogeneous Graph.} We define a heterogeneous graph as $\mathcal{G} = (\mathcal{V}, \mathcal{E}, \mathcal{A}, \mathcal{R}, \mathbf{X}, \mathcal{P})$,
where $\mathcal{V} = \{v_i\}_{i=1}^{N}$ denotes the set of $N$ nodes and
$\mathbf{X} = \{\mathbf{x}_i\}_{i=1}^{N}$ represents the corresponding node features. The edge set is defined as $\mathcal{E} = \{(v_a, r, v_b)\}_{j=1}^{M}$, where each edge connects node $v_a$ to node $v_b$ via a relation type $r \in \mathcal{R}$. $\mathcal{A}$ and $\mathcal{R}$ denote the sets of node types and relation types, respectively, with $|\mathcal{A}| + |\mathcal{R}| > 2$. $\mathcal{P} = \{p_i\}_{i=1}^{|\mathcal{P}|}$ is a set of meta-paths, where each meta-path $p_i$ is defined as a sequence of relations $r_1, r_2, \ldots, r_s$, i.e., $v_1 \xrightarrow{r_1} v_2 \xrightarrow{r_2} \cdots \xrightarrow{r_s} v_{s+1}$, and $s$ is the length of the meta-path. 

\textbf{Definition 3.2 Heterogeneous Graph Prompt Learning.}
HGPL adapts a pre-trained heterogeneous graph model to downstream applications by introducing learnable prompt parameters while keeping the pre-trained model fixed. Given a heterogeneous graph $\mathcal{G}$ with a set of meta-paths $\mathcal{P}$, HGPL injects prompts into node features or graph topology to guide task-specific learning. Formally, it can be defined as:
\begin{equation}
\tilde{\mathcal{G}}_{i} = f\big(\mathcal{G}_{i}, \mathbf{p}_i\big),
\end{equation}
where $\mathcal{G}_{i}$ denotes the graphs corresponding to meta-path $p_i \in \mathcal{P}$, $\mathbf{p}_i$ is a learnable prompt specific to $p_i$, and $f(\cdot)$ is a prompt injection function, producing the prompted graph $\tilde{\mathcal{G}}_{i}$ for downstream tasks.

\textbf{Definition 3.3 Mixture of Experts (MoE).} MoE has achieved notable success in language~\cite{MoE_for_language} and vision~\cite{MoE_for_vision} domains. It is a modular learning paradigm consisting of a set of experts and a routing function that selects and combines compatible experts. In graph representation learning, given an input representation $\mathbf{h}_i$ of a node $v_i$, an MoE model maintains an expert pool $\mathbf{E} = \{ E_1, E_2, \dots, E_\mathcal{M} \}$, where each expert $E_m(\cdot) \in \mathbb{R}^{N \times d}$ is a learnable function. A routing function $g(\cdot)$ generates the routing weights $\boldsymbol{\alpha}_i$ from  $\mathbf{h}_i$ via a softmax operation. The output of the MoE $\mathbf{h}'_i$ is obtained as:
\begin{equation}
\boldsymbol{\alpha}_i = \text{softmax}\big(g(\mathbf{h}_i)\big), \quad
\mathbf{h}'_i = \sum_{m=1}^{\mathcal{M}} \alpha_{i,m} \, E_m(\mathbf{h}_i).
\end{equation}

\section{Methodology}

\begin{figure*}[t]
    \centering
    \includegraphics[width=\textwidth,height=0.45\textheight,keepaspectratio]{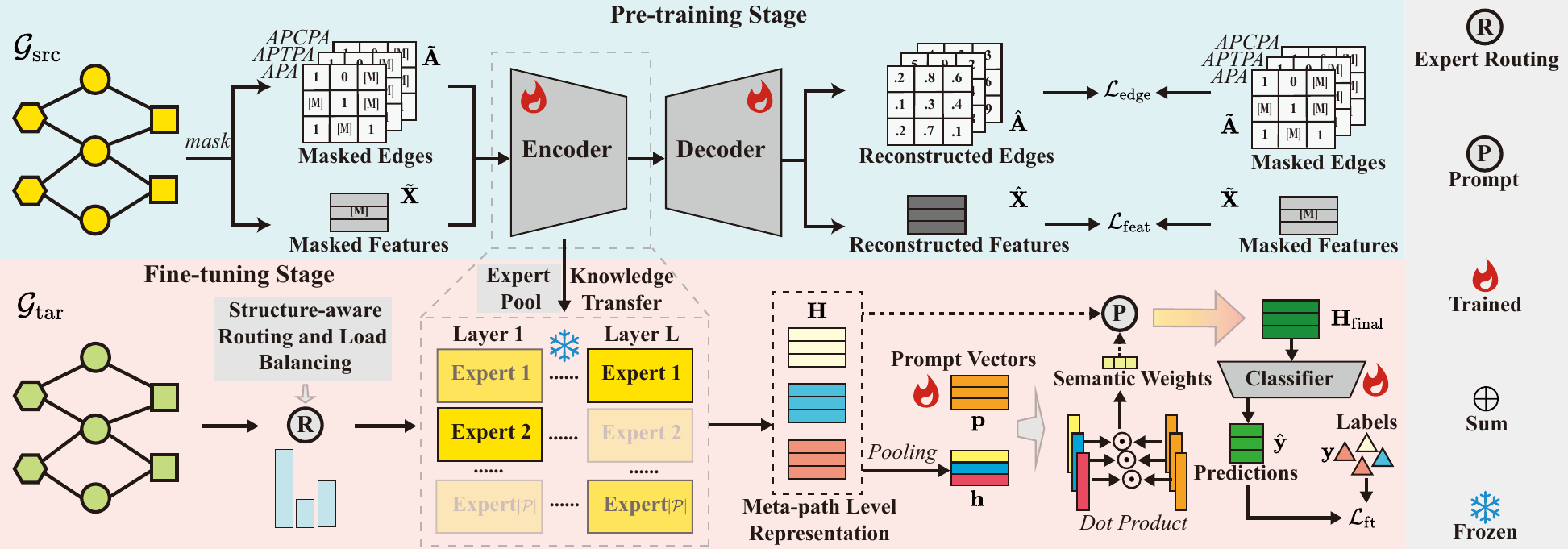}
    \caption{Overall framework of CHoE.} 
    \label{fig:model}
\end{figure*}

To resolve the challenges described in Section~\ref{sec:introduction}, we propose CHoE, which consists of four components: a self-supervised pre-training framework, a structure-conditioned experts prompt-tuning module, a structure-aware expert routing and load balancing mechanism, and a prompt-based semantic fusion module. The overall CHoE framework is illustrated in Figure~\ref{fig:model}.

\subsection{Self-supervised Graph Pre-training}
In the pre-training stage, we adopt a generative learning framework to learn general semantic and structural information in the source domain. Given a heterogeneous graph $\mathcal{G}_\text{src} = (\mathcal{V}_\text{src}, \mathbf{X}_\text{src}, \mathcal{P})$, which contains a set of pre-defined meta-paths $\mathcal{P} = \{ p_i \}_{i=1}^{|\mathcal{P}|}$, all methods in this subsection are performed on a single meta-path view unless otherwise noted. We first construct a meta-path-based adjacency matrix $\mathbf{A} \in \mathbb{R}^{N \times N}$ via meta-path sampling~\cite{Meta-path}, 
and apply Singular Value Decomposition (SVD) to align feature dimensions across domains, obtaining $N \times F$-dimensional node features $\mathbf{X}_{\text{src}} \in \mathbb{R}^{N \times F}$.
Then, we design an autoencoder architecture, in which the encoder $f_{\text{E}}(\cdot) \in \mathbb{R}^{N \times d}$ encodes node features $\mathbf{X}_\text{src}$ and meta-path-based adjacency matrices $\mathbf{A}$ to node representations, and the decoder $f_{\text{D}}(\cdot) \in \mathbb{R}^{N \times F}$ reconstructs the graph structure and node features from the learned representations.

\subsubsection{Masked Feature and Edge Reconstruction Tasks}

To achieve self-supervised heterogeneous graph pre-training, inspired by the representative masked reconstruction paradigm GraphMAE~\cite{GraphMAE}, we design masked feature and edge reconstruction tasks that aim to capture semantic and structural information in heterogeneous graphs.

In the masked feature reconstruction task, for the adjacency matrix 
$\mathbf{A}$, we construct a feature masking matrix following a Bernoulli distribution $M_\text{f} \sim \text{Bernoulli}(r_1)$, where $r_1$ is the feature masking ratio. The masked features are obtained as $\tilde{\mathbf{X}} = M_\text{f} \odot \mathbf{X}_\text{src} + (1 - M_\text{f}) \odot \mathbf{X}_{\text{[M]}}$,
where $\mathbf{X}_{\text{[M]}}$ denotes learnable mask embeddings.
We feed $\tilde{\mathbf{X}}$ and $\mathbf{A}$ into the encoder $f_{\text{E}}(\cdot)$ to obtain the node representations $\mathbf{H}_\text{f}$. Subsequently, the decoder $f_{\text{D}}(\cdot)$ 
reconstructs the node features as $\hat{\mathbf{X}} \in \mathbb{R}^{N \times F} $, which can be formulated as:

\begin{equation}
\mathbf{H}_\text{f} = 
f_{\text{E}}\!\left(\tilde{\mathbf{X}}, {\mathbf{A}}\right), 
\quad
\hat{\mathbf{X}} = 
f_{\text{D}}\!\left(\mathbf{H}_\text{f}, {\mathbf{A}}\right).
\end{equation}
The feature reconstruction loss $\mathcal{L}_\text{f}$ 
is formulated to measure the discrepancy between the $\hat{\mathbf{X}}$ and $\tilde{\mathbf{X}}$ with a scaling factor $\gamma_1$, which is defined as:

\begin{equation}
\mathcal{L}_\text{f} = \frac{1}{|\mathcal{V}_\text{src}^m|} \sum_{v \in \mathcal{V}_\text{src}^m} \left( 1 - \frac{\tilde{\mathbf{x}}_v \hat{\mathbf{x}}_v}{\|\tilde{\mathbf{x}}_v\|_2\|\hat{\mathbf{x}}_v\|_2} \right)^{\gamma_1},
\end{equation}
where $\mathcal{V}_\text{src}^m$ denotes the set of masked nodes. 
By applying the above procedure to each meta-path view $p_i \in \mathcal{P}$, we obtain $\{\mathcal{L}_\text{f}^i\}_{i=1}^{|\mathcal{P}|}$.
To aggregate $\mathcal{L}_\text{f}^{i}$ across meta-path views, we introduce an attention vector $\mathbf{q} \in \mathbb{R}^{F}$ to compute the importance score for each meta-path, and then apply a softmax to obtain the attention weight $w_\text{f}^{i}$, which is defined as: 
\begin{equation}
\label{equ:attn}
w_\text{f}^{i} = \text{softmax}\left(\mathbf{q}^{\top} \tanh\!\left(\mathbf{W}\tilde{\mathbf{X}}^i + \mathbf{b}\right)\right),
\end{equation}
where $\mathbf{W}$ is a learnable weight matrix and $\mathbf{b}$ is a bias vector. The loss $\mathcal{L}_\text{feat}$ is defined as the weighted aggregation of $\mathcal{L}_\text{f}^{\,i}$ across all meta-paths:
\begin{equation}
\mathcal{L}_{\text{feat}}
=
\sum_{i=1}^{|\mathcal{P}|}
w_{\text{f}}^i \,
\mathcal{L}_{\text{f}}^i.
\end{equation}

In the masked edge reconstruction task, for the adjacency matrix
$\mathbf{A}$, we construct an edge masking matrix $M_e \sim \text{Bernoulli}(r_2)$, where $r_2$ is the edge masking rate. The masked adjacency matrix is obtained as $\tilde{\mathbf{A}} = M_e \odot \mathbf{A}$. Given the node features $\mathbf{X}_{\text{src}} \in \mathbb{R}^{N\times F}$ and $\tilde{\mathbf{A}}$, the $f_{\text{E}}(\cdot)$ encodes the node representations $\mathbf{H}_{\text{e}}$, which are further decoded by $f_{\text{D}}(\cdot)$ to obtain $\mathbf{Z}$. This process can be formulated as:
\begin{equation}
\mathbf{H}_{\text{e}} = 
f_{\text{E}}\!\left(\mathbf{X}_\text{src}, \tilde{\mathbf{A}}\right), 
\quad
\mathbf{Z} = 
f_{\text{D}}\!\left(\mathbf{H}_{\text{e}}, \tilde{\mathbf{A}}\right).
\end{equation}
We compute the discrepancy between $\hat{\mathbf{A}} = \sigma(\mathbf{Z}{\mathbf{Z}}^{\top})$ and $\mathbf{A}$ with a scaling factor $\gamma_2$:
\begin{equation}
\mathcal{L}_{\text{e}}
=
\frac{1}{|\mathcal{V}_{\text{src}}|}
\sum_{v \in \mathcal{V}_{\text{src}}}
\left(
1 -
\frac{
\mathbf{A}_v \cdot \hat{\mathbf{A}}_v
}{
\|\mathbf{A}_v\|_2 \|\hat{\mathbf{A}}_v\|_2
}
\right)^{\gamma_2}.
\end{equation}
After applying the edge reconstruction task to all meta-paths, we obtain $\{\mathcal{L}_\text{e}^i\}_{i=1}^{|\mathcal{P}|}$.
We then calculate the corresponding attention weight $w_{\text{e}}^{i}$ according to Eq.~\ref{equ:attn}, and the loss $\mathcal{L}_{\text{edge}}$ is obtained via a weighted aggregation:
\begin{equation}
\mathcal{L}_{\text{edge}}
=
\sum_{i=1}^{|\mathcal{P}|}
w_{\text{e}}^{i} \,
\mathcal{L}_{\text{e}}^i.
\end{equation}
Finally, the pre-training loss $\mathcal{L}_{\text{pre}}$ is defined as:
\begin{equation}
\mathcal{L}_{\text{pre}}
=
\mathcal{L}_{\text{feat}}
+
\lambda_{\text{edge}}
\mathcal{L}_{\text{edge}},
\end{equation}
where $\lambda_{\text{edge}}$ is used to balance the $\mathcal{L}_{\text{feat}}$ and $\mathcal{L}_{\text{edge}}$.

\subsection{Structure-Conditioned Experts Prompt-tuning}

In the fine-tuning stage, meta-paths in downstream graphs often differ from those in the source domain in type and number, limiting the compatibility of pre-trained models with downstream graphs. To address this challenge, we introduce structure-conditioned experts.
The encoder $f_{\text{E}}(\cdot)$ has learned general knowledge in the pre-training stage. We fix its parameters and adopt it as an expert pool $E = \left\{ E^l \mid l = 1, 2 \dots L\right\}$. For the $l$-th expert $E^{l}$, it contains structure-conditioned experts, which is expressed as $E^l = \left\{ E_i^l \mid i = 1, 2 \dots |\mathcal{P}|\right\}$, where each expert corresponds to a source-domain meta-path. Each $E_i^l$ is implemented as a Graph Attention Network~\cite{GAT} encoder. 

Given a heterogeneous graph $\mathcal{G}_{\text{tar}} = (\mathcal{V}_{\text{tar}}, \mathbf{X}_{\text{tar}}, \mathcal{T})$, which contains a set of meta-paths $\mathcal{T} = \{ t_j \}_{j=1}^{|\mathcal{T}|}$, all methods in this subsection are performed on a single meta-path view unless otherwise noted.
We construct a meta-path-based adjacency matrix $\mathbf{A}$ and align feature dimensions
across domains, obtaining $N \times F$-dimensional node features $\mathbf{X}_{\text{tar}} \in \mathbb{R}^{N \times F}$. We then define $\mathbf{H}^l$ as the node representations at the $l$-th expert layer, with $\mathbf{H}^{0} = \mathbf{X}_{\text{tar}}$. Each expert $E_i^l$ generates the expert-specific node representations $\mathbf{H}_{i}^l$, which are as follows:
\begin{equation}
\mathbf{H}_i^l = E_i^l\!\left(\mathbf{A}, \mathbf{H}^{l-1}\right).
\end{equation}

\subsubsection{Structure-Aware Expert Routing and Load Balancing}
After obtaining node representations $\mathbf{H}_{i}^l$ of all experts at layer $l$, we learn routing weights to aggregate them into $\mathbf{H}^l$. Based on the structural distribution discrepancy across domains, we design a routing mechanism that evaluates the structural consistency between $\mathbf{A}$ and $\mathbf{H}_i^l$. An expert that can reconstruct a meta-path view should be assigned a large routing weight for that view. Moreover, connected nodes in the meta-path view are expected to have similar representations. Formally, given positive samples $v_a, v_b \in \mathcal{V}_{\text{tar}}$ and negative samples $v_c, v_d \in \mathcal{V}_{\text{tar}}$, we construct positive and negative sample sets:
\begin{equation}
\resizebox{\linewidth}{!}{$
\begin{aligned}
    \mathcal{E}^{+} = \left\{(v_a, v_b) \,\middle|\, \mathbf{A}_{ab} = 1 \right\}, \;\;
    \mathcal{E}^{-} = \left\{(v_c, v_d)\,\middle|\, \mathbf{A}_{cd}=0\right\},
\end{aligned}
$}
\end{equation}
where positive samples are randomly selected from connected node pairs and negative samples are sampled from other node pairs. We compute the similarity for each sampled node pair and obtain a positive score $s_{i}^{+l}$ and a negative score $s_{i}^{-l}$ by averaging the similarities within each set:
\begin{equation}
\begin{aligned}
s_{i}^{+l}
&=
\frac{1}{\left|\mathcal{E}^{+}\right|}\sum_{(v_a,v_b)\in\mathcal{E}^{+}} (\mathbf{h}_{a,i}^{l})^{\top}\mathbf{h}_{b,i}^{l}, \\
s_{i}^{-l}
&=
\frac{1}{\left|\mathcal{E}^{-}\right|}
\sum_{(v_c,v_d)\in \mathcal{E}^{-}} (\mathbf{h}_{c,i}^{l})^\top \mathbf{h}_{d,i}^{l},
\end{aligned}
\end{equation}
where $\mathbf{h}_{u,i}^{l}$ denotes the representation of node $v_u$ produced by $E_i^l$. Then, we compute the expert score $r_{i}^{l}$:
\begin{equation}
r_{i}^{l}
=
\frac{
\exp\!\left( s_{i}^{+l} / \tau \right)
}{
\exp\!\left( s_{i}^{+l} / \tau \right)
+
\exp\!\left( s_{i}^{-l} / \tau \right)
},
\end{equation}
where $\tau$ denotes the temperature parameter.

Then, we note that assigning expert weights solely based on expert scores can lead to severe expert imbalance, where a small subset of experts dominates the routing decisions across layers.
To address this issue, we further introduce a load balancing mechanism that adaptively penalizes experts based on their cumulative routing weights in previous epochs, encouraging a more balanced allocation of experts. Specifically, we maintain an expert load counter $m_i$ that records the cumulative sum of routing weights assigned to expert $E_i$ across all layers over previous epochs. At the current epoch, we compute the relative load of expert $E_i$ as:
\begin{equation}
\tilde{m}_i = \frac{m_i}{\frac{1}{|\mathcal{P}|}\sum_{k=1}^{|\mathcal{P}|} m_k + \epsilon}.
\end{equation}
Based on $\tilde{m}_i$, we use an exponential decay regularization term $\gamma_i$ to compute the final routing score $\tilde{r}^l_{i}$:
\begin{equation}
\gamma_i = \exp(-\lambda_{\text{balance}} \tilde{m}_i),
\quad
\tilde{r}^l_{i} = r^l_{i} \cdot \gamma_i,
\end{equation}
where $\lambda_\text{balance}$ controls the strength of expert balancing. Subsequently, the routing weights are computed by applying normalization to the routing scores: 
\begin{equation}
\alpha^l_{i} = 
\frac{\exp(\tilde{r}^l_{i})}
{\sum_{k=1}^{|\mathcal{P}|} \exp(\tilde{r}^l_{k})}.
\end{equation}
Finally, we aggregate these expert-specific representations $\mathbf{H}_{i}^{l}$ using the routing weights $\alpha^l_{i}$ to obtain $\mathbf{H}^{\,l}$, and update the expert load counter $m_i$ with $\alpha^l_{i}$:
\begin{equation}
\mathbf{H}^{\,l}
=
\sum_{i=1}^{|\mathcal{P}|} \alpha_{i}^{l}\, \mathbf{H}_{i}^{\,l},
\quad
m_i \gets m_i + \alpha^l_{i}.
\end{equation}

\subsubsection{Prompt-Based Semantic Fusion Module}
While the structure-aware expert routing mechanism selects the most compatible experts for each meta-path from a structural perspective, thereby producing node representations for each meta-path view, the downstream tasks further require an effective integration of these representations. Therefore, we introduce a prompt-based fusion module to adaptively aggregate representations across different semantic views.

After applying the above procedure layer by layer to each meta-path view $t_j \in \mathcal{T}$, we obtain a set of representations at the $L$-th expert layer for all meta-path views in the target domain: $\{\mathbf{H}_j^L\}_{j=1}^{|\mathcal{T}|}$. Then, to generate a global representation for each meta-path view, we perform mean pooling over all nodes: 
\begin{equation}
\mathbf{h}_j = \frac{1}{|\mathcal{V}_\text{tar}|} \sum_{v \in \mathcal{V}_\text{tar}} \mathbf{h}^{L}_{v,j},
\end{equation}
in which $\mathbf{h}_j$ denotes the global representation of meta-path $t_j$. We introduce a learnable prompt vector $\mathbf{p}_j$ for each meta-path to assess its semantic importance, computing a semantic importance score $s_j$ that is further normalized across all meta-paths to obtain the corresponding semantic weight $\beta_j$:
\begin{equation}
s_j = \mathbf{h}_j^\top \mathbf{p}_j,
\quad
\beta_j = \frac{\exp(s_j)}{\sum_{k=1}^{|\mathcal{P}|} \exp(s_{k})}.
\end{equation}
The final $\mathbf{H}_\text{final}$ is obtained by aggregating meta-path-specific node representations with their semantic weights and then fed into a linear classifier $f_{\text{C}}(\cdot)$:
\begin{equation}
\mathbf{H}_{\text{final}} = \sum_{j=1}^{|\mathcal{T}|} \beta_j \mathbf{H}_j^{L},
\quad
\hat{\mathbf{y}} = f_{\text{C}}\!\left(\mathbf{H}_{\text{final}}\right).
\end{equation}
During the fine-tuning stage, we optimize the $\mathcal{L}_{\text{ft}}$ over labeled nodes:
\begin{equation}
\mathcal{L}_{\text{ft}} = - \sum_{v \in \mathcal{V}_\text{tar}^\text{label}} \mathbf{y}_v \log \hat{\mathbf{y}}_v,
\end{equation}
where $\mathcal{V}_\text{tar}^\text{label}$ is the set of labeled nodes, $\mathbf{y}_v$ and $\hat{\mathbf{y}}_v$ denote the true label and predicted label of node $v$, respectively. Notably, only $\{\mathbf{p}_j\}_{j=1}^{|\mathcal{T}|}$, and the parameters of $f_{\text{C}}(\cdot)$, are updated during fine-tuning, enabling the pre-trained model to achieve efficient cross-domain transfer.

\section{Experiments}

\subsection{Experimental Setup}

\begin{table}[t]
\centering
\setlength{\tabcolsep}{4pt} 
\renewcommand{\arraystretch}{0.9} 

\begin{tabular}{lccc} 
\toprule
\textbf{Dataset} & \textbf{Node Type} & \textbf{Meta-path} & \textbf{Classes} \\
\midrule
\multirow{3}{*}{ACM}
& Paper (P): 4019   & \multirow{3.2}{*}{PAP, PSP} & \multirow{3}{*}{3} \\
& Author (A): 7167  &                             &  \\
& Subject (S): 60   &                             &  \\
\midrule
\multirow{4}{*}{DBLP}
& Author (A): 4057      & \multirow{4.2}{*}{\shortstack{APA, APCPA, \\ APTPA}} & \multirow{4}{*}{4} \\
& Paper (P): 14328      &                                                      &  \\
& Conference (C): 20    &                                                      &  \\
& Term (T): 7723        &                                                      &  \\
\midrule
\multirow{4}{*}{Freebase}
& Movie (M): 3492   & \multirow{4.2}{*}{\shortstack{MAM, MDM, \\ MWM}} & \multirow{4}{*}{4} \\
& Actor (A): 33401  &                                                  &  \\
& Director (D): 2502 &                                                 &  \\
& Writer (W): 4459  &                                                  &  \\
\midrule
\multirow{3}{*}{AMiner}
& Paper (P): 6564     & \multirow{3.2}{*}{PAP, PRP} & \multirow{3}{*}{3} \\
& Author (A): 13329   &                             &  \\
& Reference (R): 35890 &                             &  \\
\bottomrule
\end{tabular}
\caption{Statistics of heterogeneous graph datasets.}
\label{tab:dataset_statistics}
\end{table}

\begin{table*}[!ht]
\centering

\fontsize{9}{11}\selectfont
\setlength{\tabcolsep}{3.5pt}

\begin{tabular}{lcccccccc}
\toprule
\multirow{2}{*}{Method} 
& \multicolumn{2}{c}{ACM} 
& \multicolumn{2}{c}{DBLP} 
& \multicolumn{2}{c}{Aminer} 
& \multicolumn{2}{c}{Freebase} \\

& Macro-F1 & Micro-F1 
& Macro-F1 & Micro-F1 
& Macro-F1 & Micro-F1 
& Macro-F1 & Micro-F1 \\
\midrule
DGI        & 81.85$\pm$4.95 & 81.73$\pm$5.44
           & 78.55$\pm$6.86 & 79.74$\pm$5.48
           & 25.76$\pm$3.45 & 32.35$\pm$6.85
           & 32.73$\pm$1.97 & 34.89$\pm$2.72 \\
GRACE      & 82.55$\pm$6.17 & 82.01$\pm$6.92
           & 70.60$\pm$5.74 & 71.78$\pm$5.80
           & 24.64$\pm$2.50 & 29.83$\pm$5.31
           & 33.08$\pm$2.72 & 35.14$\pm$3.78 \\
\midrule
HeCo       & 74.40$\pm$3.80 & 76.37$\pm$2.19
           & 87.27$\pm$0.89 & 88.06$\pm$0.90
           & 24.07$\pm$2.34 & 27.64$\pm$3.97
           & 36.68$\pm$3.69 & 40.07$\pm$3.64 \\
HGMAE      & 81.32$\pm$4.65 & 81.31$\pm$4.32
           & 75.44$\pm$8.47 & 76.75$\pm$7.93
           & 24.27$\pm$3.43 & 31.66$\pm$6.90
           & 32.45$\pm$2.36 & 35.79$\pm$3.47 \\
HERO       & \underline{83.24$\pm$7.86} & \underline{83.83$\pm$7.32}
           & 80.40$\pm$5.70 & 80.75$\pm$5.65
           & 27.98$\pm$8.23 & 29.50$\pm$7.81
           & 31.41$\pm$7.89 & 33.33$\pm$7.30 \\
\midrule
GraphPrompt & 70.64$\pm$6.91 & 70.19$\pm$8.87
            & 87.03$\pm$0.81 & 87.82$\pm$0.79
            & 28.04$\pm$2.04 & 31.52$\pm$2.86
            & 39.96$\pm$3.18 & 43.39$\pm$2.97 \\
GPF-plus   & 82.20$\pm$7.83 & 82.67$\pm$8.08
           & 56.45$\pm$14.24 & 59.12$\pm$12.21
           & 20.27$\pm$4.38 & 33.57$\pm$11.30
           & 28.43$\pm$3.75 & 36.61$\pm$5.40\\
EdgePrompt & 33.72$\pm$17.05 & 46.80$\pm$16.02
           & 56.67$\pm$13.31 & 61.22$\pm$11.65
           & 15.77$\pm$4.75 & 32.92$\pm$15.78
           & 23.72$\pm$4.16 & 34.37$\pm$7.23 \\
\midrule
HGPrompt   & 71.36$\pm$6.68 & 71.08$\pm$7.96
           & \underline{88.18$\pm$2.15} & \underline{89.01$\pm$2.00}
           & \underline{29.56$\pm$2.64} & 35.81$\pm$4.12
           & \underline{40.91$\pm$3.26} & 42.83$\pm$4.06 \\
HetGPT     & 65.56$\pm$8.14 & 72.74$\pm$3.94
           & 58.83$\pm$17.16 & 64.13$\pm$14.19
           & 20.74$\pm$2.67 & \underline{48.01$\pm$8.40}
           & 29.94$\pm$7.44 & \underline{45.65$\pm$2.21} \\
\midrule
GraphLoRA  & 67.86$\pm$11.90 & 69.47$\pm$9.79
           & 69.21$\pm$9.75 & 70.43$\pm$9.71
           & 25.29$\pm$1.97 & 28.68$\pm$2.19
           & 32.92$\pm$1.22 & 34.08$\pm$1.32 \\
\midrule
\textbf{CHoE} & \textbf{83.27$\pm$2.19} & \textbf{83.79$\pm$1.92}
           & \textbf{88.47$\pm$0.90} & \textbf{89.47$\pm$0.75}
           & \textbf{53.87$\pm$6.41} & \textbf{62.16$\pm$6.99}
           & \textbf{49.75$\pm$4.38} & \textbf{51.81$\pm$5.47} \\
\bottomrule
\end{tabular}
\caption{Cross-domain node classification. Macro-F1 and Micro-F1 for 5-shot tasks, where all models are pre-trained on the ACM dataset and fine-tuned on four target datasets, over eleven baselines. The best results are highlighted in bold, and the runner-up with an underline.}
\label{tab:main_results}
\end{table*}

\begin{figure*}[!ht]
    \centering    \includegraphics[width=\textwidth,height=0.45\textheight,keepaspectratio]{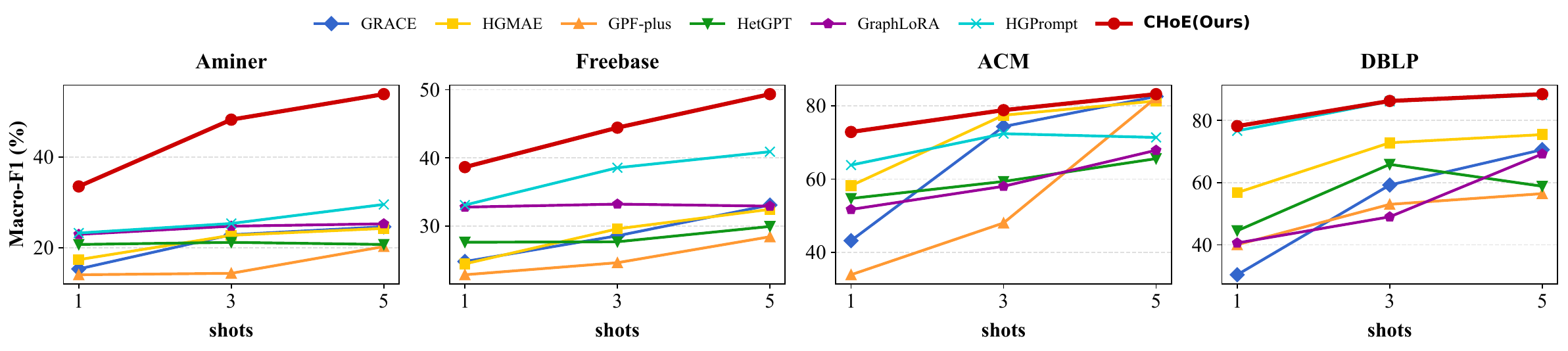}
    \caption{Cross-domain node classification over four datasets under different shot settings.}
    \label{fig: cmp_shots}
\end{figure*}

\subsubsection{Datasets and Baselines}
In our experiments, we adopt four widely used heterogeneous graph datasets: ACM, DBLP, Aminer, and Freebase. Each dataset consists of multiple node types and meta-paths. The statistics of the datasets are shown in Table~\ref{tab:dataset_statistics}.
We compare our method with five representative categories of existing methods, including:
(1) homogeneous self-supervised methods: DGI~\cite{DGI} and GRACE~\cite{Grace}; 
(2) heterogeneous self-supervised methods: HeCo~\cite{HeCo}, HGMAE~\cite{HGMAE}, and HERO~\cite{HERO}; 
(3) homogeneous graph prompt learning methods: GraphPrompt~\cite{GraphPrompt}, GPF-plus~\cite{GPF-plus}, and EdgePrompt~\cite{EdgePrompt};
(4) heterogeneous graph prompt learning methods: HGPrompt~\cite{HGPrompt} and HetGPT~\cite{HetGpt};
(5) cross-domain homogeneous graph prompt learning methods: GraphLoRA~\cite{GraphLoRA}.

\subsubsection{Implementation Details}
We follow~\cite{Meta-path} to select meta-paths for each dataset. For the proposed CHoE, we adopt HAN~\cite{HAN} as the encoder and decoder in the pre-training stage. For both the baselines and our method, models are pre-trained for 10,000 epochs and fine-tuned for 500 epochs, with early stopping (patience=20). We independently searched for the learning rates for the pre-training and fine-tuning stages within the range of 1e-5 to 0.01. To obtain reliable results, we perform 20 repeated runs for each of 5 fixed random seeds, reporting average results over 100 trials.

\subsection{Cross-Domain Node Classification}

\textbf{5-shot experiments on different downstream datasets.} 
We report 5-shot node classification on four datasets using ACM for pre-training. As shown in Table~\ref{tab:main_results}, CHoE consistently outperforms all baselines, with the largest gains on Aminer and Freebase. On DBLP, the improvement is smaller because its dense graph structure enables baselines (e.g., HGPrompt and GraphPrompt) to capture sufficient structural information during fine-tuning, reducing the difficulty of cross-domain transfer. On ACM, CHoE achieves competitive performance against strong in-domain baselines (e.g., HERO and GRACE), demonstrating its effectiveness in both cross-domain and in-domain settings.

\begin{table*}[t]
\centering
\small
\begin{tabular}{cllllllll}
\toprule
\multirow{2}{*}{Variants} 
& \multicolumn{2}{c}{ACM} 
& \multicolumn{2}{c}{DBLP} 
& \multicolumn{2}{c}{Aminer} 
& \multicolumn{2}{c}{Freebase} \\
& Macro-F1 & Micro-F1 
& Macro-F1 & Micro-F1 
& Macro-F1 & Micro-F1 
& Macro-F1 & Micro-F1 \\
\midrule
CHoE 
& \textbf{83.27$\pm$2.19} & \textbf{83.79$\pm$1.92} 
& \textbf{88.47$\pm$0.90} & \textbf{89.47$\pm$0.75} 
& \textbf{53.87$\pm$6.41} & \textbf{62.16$\pm$6.99} 
& \textbf{49.75$\pm$4.38} & \textbf{51.81$\pm$5.47} \\
w/o PR 
& 75.21$\pm$4.07 & 75.20$\pm$4.80 
& 83.67$\pm$4.58 & 84.90$\pm$4.00 
& 48.14$\pm$7.11 & 56.95$\pm$5.69 
& 46.41$\pm$5.43 & 48.73$\pm$5.92 \\
w/o LB 
& 80.36$\pm$3.41 & 81.12$\pm$3.42 
& 84.31$\pm$13.85 & 84.94$\pm$12.99 
& 45.82$\pm$10.52 & 57.12$\pm$8.80 
& 46.27$\pm$6.50 & 48.34$\pm$7.16 \\
w/o ER 
& 71.16$\pm$9.55 & 75.17$\pm$7.97 
& 74.16$\pm$5.84 & 75.45$\pm$5.33 
& 42.13$\pm$8.60 & 49.43$\pm$9.02 
& 43.99$\pm$5.19 & 45.77$\pm$6.08 \\

\bottomrule
\end{tabular}
\caption{Ablation study. ``w/o PR'', ``w/o LB'', and ``w/o ER'' denote variants without the prompt module, load balancing mechanism, and structure-aware expert routing mechanism, respectively.}
\label{tab:ablation}
\end{table*}

\begin{table}[t]
\centering
\setlength{\tabcolsep}{2pt}
\small
\begin{tabular}{lccccc}
\toprule
\multirow{2}{*}{Upstream} 
& \multicolumn{4}{c}{Macro-F1} 
& \multirow{2}{*}{Downstream} \\

& HGMAE & HGPrompt & HetGPT & CHoE & \\
\midrule

\multirow{4}{*}{ACM}
& 81.32 & 71.36 & 65.56 & \textbf{83.18} & ACM \\
& 75.44 & 88.18 & 58.83 & \textbf{88.45} & DBLP \\
& 24.27 & 29.56 & 20.74 & \textbf{53.87} & Aminer \\
& 32.45 & 40.91 & 29.94 & \textbf{49.36} & Freebase \\
\midrule

\multirow{4}{*}{DBLP}
& 80.10 & 72.88 & 58.49 & \textbf{79.63} & ACM \\
& 72.92 & \textbf{85.68} & 73.80 & 76.38 & DBLP \\
& 26.22 & 26.77 & 21.26 & \textbf{47.41} & Aminer \\
& 30.62 & 38.97 & 30.06 & \textbf{46.19} & Freebase \\
\midrule

\multirow{4}{*}{Aminer}
& 78.32 & 71.92 & 51.53 & \textbf{79.63} & ACM \\
& 61.16 & \textbf{82.29} & 67.68 & 81.74 & DBLP \\
& 27.66 & 31.17 & 19.79 & \textbf{53.43} & Aminer \\
& 32.25 & 39.36 & 30.09 & \textbf{46.16} & Freebase \\
\midrule

\multirow{4}{*}{Freebase}
& 78.34 & 71.14 & 45.71 & \textbf{80.58} & ACM \\
& 60.07 & 81.90 & 60.97 & \textbf{84.30} & DBLP \\
& 26.93 & 26.85 & 19.81 & \textbf{54.39} & Aminer \\
& 31.51 & 42.39 & 32.13 & \textbf{48.58} & Freebase \\

\bottomrule
\end{tabular}
\caption{Cross-domain node classification. Macro-F1 on 5-shot cross-domain node classification tasks under all combinations of upstream and downstream datasets from four datasets.}
\label{tab:all_different}
\end{table}

\textbf{1/3-shot experiments on different downstream datasets.} 
We conduct 1-shot and 3-shot node classification experiments using ACM for pre-training to evaluate the cross-domain adaptability of CHoE. As shown in Figure~\ref{fig: cmp_shots}, CHoE consistently outperforms all baselines across different settings. The performance gains are more significant on Aminer and Freebase as the number of shots increases, while stable improvements are observed on DBLP. On ACM, CHoE remains competitive with strong in-domain baselines.

\textbf{5-shot experiments on different pre-training datasets.} 
To further evaluate the cross-domain capability of CHoE, we conduct 5-shot node classification on all combinations of ACM, DBLP, Aminer, and Freebase as upstream and downstream datasets, as shown in Table~\ref{tab:all_different}. CHoE achieves superior and more stable performance across most settings. Although HGPrompt shows higher performance in some upstream domains, its results often performance under domain shifts, indicating limited transferability. In contrast, CHoE generalizes better across domains, particularly on Aminer and Freebase, demonstrating stronger cross-domain knowledge transfer ability.

\subsection{Ablation Study}
To evaluate the effectiveness of each component in CHoE, we compare it with three variants: ``w/o PR'' (without the prompt module), ``w/o LB'' (without load balancing), and ``w/o ER'' (without structure-aware expert routing and load balancing). As shown in Table~\ref{tab:ablation}, we report 5-shot node classification on four datasets using ACM for pre-training. Removing expert routing leads to the largest performance drop, highlighting the importance of selecting structure-compatible experts for downstream graphs. Performance also decreases for w/o PR'' and ``w/o LB'', demonstrating the effectiveness of the prompt module and load balancing mechanism.

\subsection{Hyperparameter Analysis}
\begin{figure}[t]
    \centering
    \includegraphics[width=\linewidth,height=0.45\textheight,keepaspectratio]{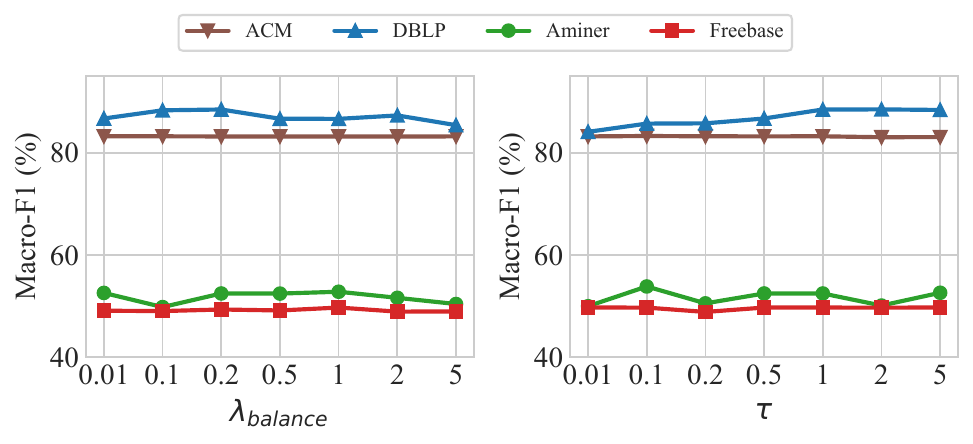}
    \caption{Hyperparameter analysis. Sensitivity of the balancing coefficient $\lambda_{\text{balance}}$ and the temperature parameter $\tau$ on four datasets.}
    \label{fig:hyper}
\end{figure}

To evaluate the sensitivity of CHoE to hyperparameters, we conduct analyses on the balancing coefficient $\lambda_{\text{balance}}$, the temperature parameter $\tau$, and the number of model layers $L$. As shown in Figure~\ref{fig:hyper}, the performance remains stable across a wide range of $\lambda_{\text{balance}}$ and $\tau$, indicating that CHoE is not sensitive to these hyperparameters. We further investigate the effect of the model depth by varying the number of layers $L$. As shown in Table~\ref{tab:model_layer}, CHoE achieves the best performance when $L=2$ on all four datasets. 1-layer model lacks sufficient capacity to capture the structural and semantic knowledge from source domain. Increasing the number of layers to 3 leads to performance degradation, which may be caused by over-smoothing and the introduction of redundant information. This result suggests that a moderate model depth is more effective.   

\begin{table}[t]
\centering
\begin{tabular}{lccccc}
\toprule
$L$ & ACM   & DBLP  & Aminer & Freebase \\ \midrule
1           & 80.45 & 79.10 & 46.45  & 35.36    \\
2           & \textbf{83.27} & \textbf{88.47} & \textbf{53.87}  & \textbf{49.75} \\
3           & 78.69 & 82.55 & 45.94  & 35.13    \\ \bottomrule
\end{tabular}
\caption{Hyperparameter analysis. Macro-F1 of different model layers $L$ on four datasets.}
\label{tab:model_layer}
\end{table}

\section{Conclusion}
In this work, we find that existing HGPL methods rely on an in-domain assumption. Through a motivation experiment, we observe that HGPL methods suffer from severe performance degradation under domain shifts. Based on the observation, we identify two key challenges: model incompatibility and knowledge transfer difficulties. To address these challenges, we propose CHoE. CHoE introduces structure-conditioned experts to transfer a pre-trained expert pool to downstream domains. It further incorporates a structure-aware expert routing and load balancing mechanism to select structurally compatible experts, and a prompt-based semantic fusion module to integrate different semantic views. Extensive experiments validate the effectiveness of CHoE in cross-domain and in-domain scenarios. In summary, we propose a cross-domain HGPL method that accounts for model compatibility and cross-domain knowledge transfer, offering new insights for the heterogeneous graph foundation models.

\appendix

\section*{Acknowledgments}
This work was supported by the National Natural Science Foundation of China (No. 62422210, No. 62276187, No. 92370111, and No. 62272340), the Hong Kong RGC Theme-based Strategic Target Grant Scheme (STG STG1/M-501/23-N), the Hong Kong Global STEM Professor Scheme, and the Hong Kong Jockey Club Charities Trust.

\bibliographystyle{named}
\bibliography{ijcai26}
\end{document}